%% file: main_arxiv.tex
\documentclass[preprint,12pt]{elsarticle}

\usepackage[a4paper, total={6in, 8in}]{geometry}
\usepackage{lmodern}

\usepackage{amssymb}
\usepackage{amsmath}
\usepackage{url}
\usepackage{xcolor}
\usepackage{color, colortbl}
\definecolor{Gray}{gray}{0.9}

\usepackage{subfigure}

\usepackage{makecell}
\usepackage{multirow}
\usepackage{diagbox}
\usepackage{enumitem}
\usepackage{tabularx, ragged2e}
\usepackage{float}
\usepackage[ruled,vlined]{algorithm2e}
\usepackage[hidelinks]{hyperref}

\begin{document}

\begin{frontmatter}



\title{Toward an Energy-Optimized Operation of Data Centers Located in Wind Farms Using Reinforcement Learning}


\author[Paderborn]{Jan Stenner} 
\author[Passau]{Alexander Kilian}
\author[Dortmund]{Sebastian Peitz}
\author[Passau]{Hermann de Meer}
\affiliation[Paderborn]{organization={Software Innovation Campus Paderborn, Paderborn University},
            country={Germany}}
\affiliation[Passau]{organization={Chair of Computer Networks and Computer Communications, University of Passau},
            country={Germany}}
\affiliation[Dortmund]{organization={Lamarr Institute for Machine Learning and Artificial Intelligence, Technical University of Dortmund},
            country={Germany}}

\begin{abstract}
This paper studies Reinforcement Learning as an online controller for curtailment-aware workload shifting in wind-turbine-integrated high-performance computing (HPC) data centers. We introduce a reproducible fixed-day simulation framework with synthetic wind and price signals and delayed completion feedback, designed to be extensible toward more complex scenarios. As a controlled benchmarking basis, we then focus on the minimal case with one wind turbine and one co-located data center. In this setting, pure Reinforcement Learning exhibits a pronounced credit-assignment problem and tends to underuse free wind energy early in the day. We therefore evaluate two complementary countermeasures: optimization-based Imitation Learning and potential-based Reward Shaping. Across multi-seed training and a 200-day test set, Proximal Policy Optimization (PPO) and a Soft Actor-Critic (SAC) variant with an additional on-policy update routine achieve strong empirical performance among learned policies, and both Imitation Learning and Reward Shaping provide improvements in relevant configurations. A performance gap to the optimizer remains, which is expected: the optimizer plans offline with full-day foresight, whereas Reinforcement Learning must decide online from current observations without future realizations. The benchmark and ablation results provide a transparent basis for extending the approach toward richer multi-site and continuous-time scenarios.
\end{abstract}




\end{frontmatter}



\input{1_Introduction}

\input{2_Related_Work}
\input{3_Background_Motivation}
\input{4_Problem_Formulation}
\input{5_Fixed_Day_Minimal}
\input{8_Conclusion}

\section*{Acknowledgment}
This work was funded by the German Federal Ministry of Research, Technology and Space (BMFTR), under grant agreement no. 16ME0619 and no. 16ME0617K, within the "Energieoptimierte Supercomputer-Netzwerke durch die Nutzung von Windenergie" (ESN4NW) project. \\

\section*{Declaration of generative AI and AI-assisted technologies in the manuscript preparation process}

During the preparation of this work, the authors used OpenAI Codex to improve the language and readability of parts of the manuscript. The tool was not used to generate research ideas, perform analyses, create figures, or draw scientific conclusions. After using this tool, the authors reviewed and edited the content as needed and take full responsibility for the content of the published article.

\bibliographystyle{elsarticle-num}
\bibliography{references_RL}

\clearpage
\appendix
\input{A_Appendix_SAC_OnPolicy}
\input{B_Appendix_PerEpisode_Metrics}

\end{document}

%% file: 1_Introduction.tex
\section{Introduction}
\label{sec:introduction}
International Energy Agency (IEA) reporting places Data Center (DC) electricity use in 2022 at roughly 1--1.3\% of global demand. Over the 2015--2022 period, the same source indicates a substantial increase in sectoral energy use \cite{IEA2023a}. At the same time, decarbonization pathways require substantial emission reductions from digital infrastructure by 2030 \cite{IEA2023a}. In parallel, power systems are shifting toward low-emission generation: renewable electricity is projected to rise strongly, and the combined share of wind and solar is expected to increase from 12\% in 2022 to 58\% by 2035 \cite{IEA2023b}. This development strengthens the operational relevance of flexible demand, because renewable integration is increasingly constrained by temporal mismatch and curtailment \cite{Chien2024}.

For computing systems, this motivates control strategies that align workload execution with renewable availability. Prior work shows that geographically and temporally flexible computing demand can absorb curtailed renewable generation and reduce emissions \cite{Zheng2020}. More broadly, stranded renewable power has also been discussed as an opportunity for large-scale and high-performance computing (HPC) infrastructures \cite{Yang2018}. However, local DC adaptation without sufficient system-level coordination can provide limited benefits and may induce adverse grid-side effects \cite{Lin2023}.

A common response is workload shifting. Production evidence confirms that temporally flexible workloads can be delayed toward lower-carbon hours at scale \cite{Radovanovic2023}. Yet this paradigm has practical limits: large-scale analyses report constrained realizable gains under real workload and system constraints \cite{Sukprasert2024}. In addition, delay-intolerant workloads challenge purely temporal shifting concepts and motivate complementary control designs \cite{Hewage2025}.

Recent studies are devoted to the energy-optimized operation of a distributed HPC infrastructure that is integrated into wind turbine environments \cite{Kilian25, Kilian2026, Ahmadi2025}. These articles aim to find a proper trade-off between the main concerns \textit{sustainability} and \textit{performance} for the wind turbine-housed DC infrastructure. For instance, \cite{Kilian2026} proposes a workload execution approach that allows for workload shifts of non-time-critical compute jobs, but also handles delay-intolerant workloads in a timely manner.
A major advantage of this novel DC infrastructure is the reduction of power curtailment of wind farms. Due to the volatile nature of renewable energy sources, maintaining the balance between generation and demand and avoiding line overloading becomes increasingly harder. For the sake of grid stability, the responsible grid operator can therefore influence the power output of a wind farm by adjusting it in accordance with a predetermined setpoint. Excess wind power beyond this setpoint is curtailed, and the resulting curtailment energy cannot be used. However, by deploying compute nodes within a wind farm collector grid, otherwise curtailed wind energy can be used for DC supply, thereby improving the effective power output of wind turbines and improving operational sustainability of DCs.

As addressed in \cite{Kilian25, Kilian2026}, the consideration of data-driven approaches such as Reinforcement Learning (RL), which is capable of real-time decision making having regard to uncertainties such as forecast errors and short-term changes of reference signals, is a worthwhile topic to be tackled. Using artificial intelligence in the context of operational sustainability can simultaneously improve sustainability of compute-extensive workloads, such as artificial intelligence.

This paper focuses on a deliberately minimal but structurally expressive fixed-day problem setting: one wind turbine with one co-located HPC-DC. The setting is intentionally hand-modeled and benchmarkable. It preserves core scheduling difficulties under volatile renewable supply while avoiding confounding factors from larger architectures.

Our central question is whether RL is a suitable online controller for workload shifts and load distribution in this setting. During development, we observe a persistent training issue: agents underuse early-episode free wind energy, even though this energy would otherwise be curtailed. We interpret this as a credit-assignment problem caused by delayed completion signals. We therefore study two countermeasures in a controlled evaluation: optimization-based Imitation Learning (IL) and potential-based Reward Shaping (RS).

The main contributions of this work are the following:
\begin{enumerate}
    \item We formalize a fixed-day RL environment for HPC-DCs housed in wind turbines with explicit step dynamics, curtailment-aware energy accounting, and a derivative-based observation design that explicitly includes free wind energy.
    \item We formulate and implement a discrete optimization benchmark as an independent reference for the fixed-day task. This benchmark serves two roles: a method-agnostic comparison point and a source of expert trajectories for optimization-based IL.
    \item We evaluate two explicit mitigation strategies for delayed-feedback credit assignment, namely optimization-based IL and potential-based RS, in a reproducible ablation pipeline with controlled synthetic signals and fixed validation/test protocols. The code for all experiments reported in this paper is publicly available at \url{https://github.com/janstenner/ESN4NW-RL-public}.
\end{enumerate}
These contributions establish a controlled methodological baseline for subsequent data-driven and continuous-operation extensions. \\ [1ex]

The remainder of this article is organized as follows: Section \ref{sec:related-work} addresses related work, particularly RL-based approaches in computing and smart grid control as well as non-RL work on curtailment-aware computing. In Section \ref{sec:background-motivation}, we provide background on RL principles to motivate the chosen control perspective. Section \ref{sec:problem-formulation-assumptions} introduces the problem formulation and assumptions. Section \ref{sec:fixed-day-minimal} presents the fixed-day minimal environment, methods, and results. Section \ref{sec:conclusion-future-work} concludes the article. \\[1ex]

%% file: 2_Related_Work.tex
\section{Related Work}
\label{sec:related-work}

Related work can be grouped into two strands relevant to this paper: non-RL research on curtailment-aware and sustainable computing, and RL-based control and scheduling methods for computing and energy systems.\\ [1ex]

\textbf{Curtailment and Sustainable Computing (without RL):}\\
Recent studies show that flexible computing demand can support renewable integration, but with important structural constraints. Workload migration between DCs can reduce curtailment and emissions when unused cross-site capacity is available \cite{Zheng2020}, and stranded renewable power has been discussed as an opportunity for large-scale and HPC-oriented computing \cite{Yang2018}. At grid scale, persistent curtailment growth further highlights the need for stronger flexible-load coupling mechanisms \cite{Chien2024}, while the coordination scope remains critical because local adaptation alone can underperform broader planning approaches \cite{Lin2023}.

As discussed in Section \ref{sec:introduction}, a novel approach that seeks to effectively reduce curtailment and to improve sustainability of computing investigates the deployment of HPC-DCs in wind turbines and the distribution of such an HPC infrastructure across several wind farms \cite{Kilian25, Kilian2026, Ahmadi2025}. In \cite{Kilian25}, a general overview and selected challenges are provided. In \cite{Kilian2026, Ahmadi2025}, moving-horizon job scheduling algorithms for DCs located in wind farms are proposed. These works seek to improve usage of curtailment energy and to reduce economic and ecological costs of DC operation.  While these algorithms are based on classical, discrete optimization under explicit completion constraints, our paper studies an RL-based, sequential online control of one co-located HPC node that also seeks to unite performance and operational sustainability, particularly by using curtailment energy for DC supply. \\ [1ex]

\textbf{RL and Computing:}\\
Recent work shows that RL and related deep-learning methods improve scheduling and resource management across edge-cloud, fog, Internet of Things (IoT), and cloud systems by balancing latency, execution time, energy use, and cost \cite{Jayanetti2022, Swarup2021, Shadroo2021, Oudaa2021, Zhou2022}. Beyond direct task scheduling, RL is also used for algorithm selection and infrastructure control, including virtual machine consolidation and network-level bandwidth/routing optimization, with reductions in power consumption and service violations \cite{Zhou2022, Shaw2022, Wang2022}. For HPC specifically, RLBackfilling shows that learned policies can outperform static backfilling heuristics under uncertain workloads \cite{KolkerHicks2023}. Relative to our focus, these works emphasize compute-side efficiency and queueing performance, whereas we center curtailment-aware online load allocation in a wind-coupled HPC benchmark and analyze delayed-feedback credit assignment explicitly. \\ [1ex]

\textbf{RL and Smart Grids:}\\
RL in smart energy systems covers microgrid battery scheduling under stochastic photovoltaics/load dynamics \cite{Leo2014, Muriithi2021}, integrated electricity-gas operation with continuous-action control \cite{Zhang2020}, and renewable-accommodation analysis under uncertainty-aware Deep Reinforcement Learning (DRL) pipelines \cite{Liu2020}. Survey and review studies place these contributions in a broader operation, control, and market context for sustainable energy systems \cite{Yang2020, Ahmad2022}. The methodological overlap with our study is the sequential decision setting under uncertainty. The key difference is the controlled wind-HPC fixed-day task, where the main objective is demand-side allocation with explicit completion requirements and targeted mitigation of delayed-reward learning effects (IL and RS). \\ [1ex]

Taken together, prior work motivates renewable-aware control but does not evaluate online RL control of time-constrained scenarios with an explicit IL/RS analysis of delayed credit assignment in early free-energy utilization.

%% file: 3_Background_Motivation.tex
\section{Background: Reinforcement Learning}
\label{sec:background-motivation}

\begin{figure}[!ht]
    \centering
    \includegraphics[width=0.7\linewidth]{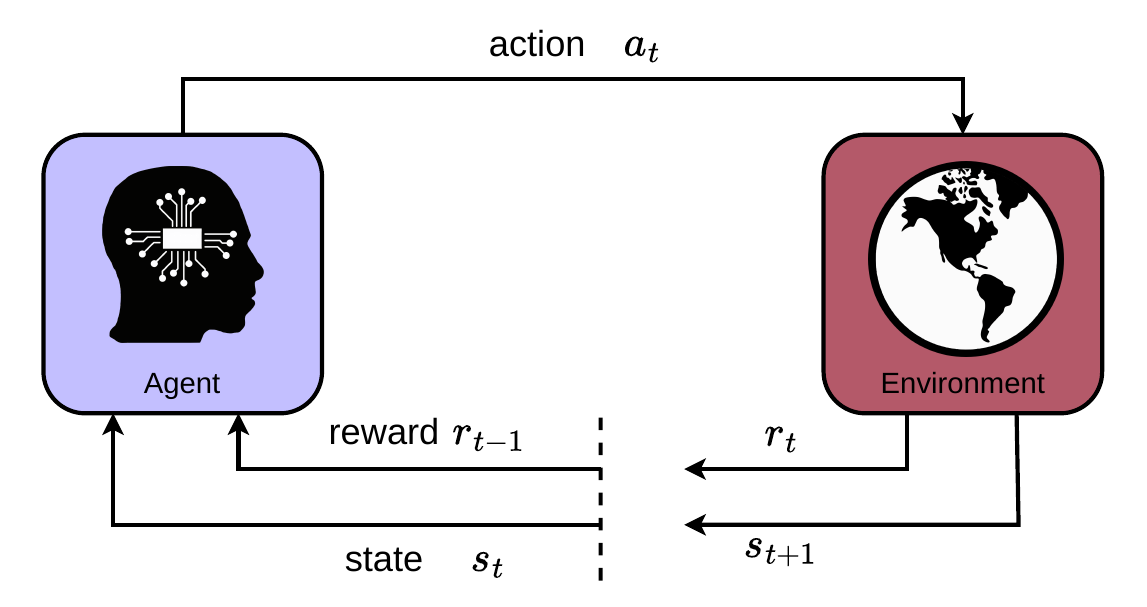}
    \caption{The agent-environment interaction in RL.}
    \label{fig:RL}
\end{figure}

In the following, we summarize the fundamentals of RL on which the remainder of this paper builds. A more detailed introduction with further explanations of concepts and methods can be found in \cite{SB18}. The general idea of RL is schematically illustrated in Figure \ref{fig:RL}: At each time instant $t$, an agent has access to system state information of the environment $s_t$, commonly referred to simply as the state. The agent interacts with the environment to achieve a predetermined goal, usually maximizing the expected cumulative reward. After deciding on a certain action $a_t$ out of a pool of feasible actions, the environment changes accordingly and presents the agent the new system state $s_{t+1}$. Furthermore, it provides a reward $r_t$ which acts as a feedback signal for the agent.

Formally, the environment is modeled as a Markov Decision Process (MDP), defined by the tuple $\big(  \mathcal{S}, \mathcal{A}, P, R, \gamma\big)$. $\mathcal{S}$ is the set of all possible states, $\mathcal{A}$ is the set of all possible actions, $P(s' \mid s,a)$ is the state transition probability function giving the likelihood of moving to state $s'$ when taking action $a$ in state $s$, $R(s,a,s')$ is the reward function yielding the reward after picking the action $a$ in state $s$ and ending up in state $s'$, and $\gamma \in (0,1]$ is a discount factor that quantifies the importance of future rewards.

For the experiments in this paper, the relationship between the environment and the information available to the agent is more accurately captured by a \emph{Partially Observable Markov Decision Process} (POMDP, \cite{Spaan2012}), given by $\big(  \mathcal{S}, \mathcal{O}, \mathcal{A}, P, \Omega, R, \gamma\big)$. Here, $\mathcal{O}$ denotes the observation space and $\Omega(o \mid s)$ is the observation function that maps the latent environment state to an observation available to the agent. To keep notation compact in the following, we continue to use the MDP-style notation and write $s_t$ for the information available to the agent at time $t$.

The agent is trying to find a policy $\pi(a \mid s)$, which is a mapping that states the likelihood of taking action $a$ provided the system state is given by $s$, so that following this policy maximizes the expected cumulative discounted reward over time $\mathbb{E}_{\pi} \left[\sum_{\tau=t}^{\infty} \gamma^{\tau}r_{\tau}\right]$ for every $t$. Two additional core concepts of RL are the value function $V$ and the state-value or Q function $Q$ which, in their recursive forms, are defined by
\begin{align*}
V^\pi(s)
&= \sum_{a\in\mathcal{A}}\pi(a \mid s)\sum_{s'\in\mathcal{S}}P(s'\mid s,a)\Bigl[R(s,a,s') + \gamma V^\pi(s')\Bigr],\\
Q^\pi(s,a)
&= \sum_{s'\in\mathcal{S}}P(s'\mid s,a)\Bigl[R(s,a,s') + \gamma \sum_{a'\in\mathcal{A}}\pi(a'\mid s')\,Q^\pi(s',a')\Bigr].
\end{align*}
The value function can be interpreted as an estimator for the expected return of following policy $\pi$ from state $s$ onward while the Q function estimates the expected return following action $a$ in state $s$ and following $\pi$ after that. In recent years, the development of DRL has led to methods, that approximate value functions and policies with neural network-based models. These approaches are especially effective when state and action spaces become large or continuous.

In this work, we consider three actor-critic DRL algorithms for continuous control: Proximal Policy Optimization (PPO, \cite{Schulman2017PPO}), Deep Deterministic Policy Gradient (DDPG, \cite{Lillicrap2016DDPG}) and Soft Actor-Critic (SAC, \cite{Haarnoja2018SAC}). PPO is an on-policy method and updates the policy from trajectories collected with the current policy only. DDPG and SAC are off-policy methods and therefore learn from replay-buffer samples that may stem from older behavior policies. This typically improves sample efficiency, while on-policy updates are often more stable but require more fresh interaction data.

In our experiments, we further use a custom SAC variant that augments the standard off-policy SAC updates with an additional on-policy update routine. The motivation is the episode-level credit-assignment structure of our fixed-day task: especially in IL settings and in runs without RS, meaningful action evaluation depends on coherent full-trajectory context because early decisions are linked to delayed completion signals. The hybrid design therefore combines replay-based sample efficiency with periodic trajectory-consistent refinement on recent rollouts.

%% file: 4_Problem_Formulation.tex
\section{Problem Formulation and Assumptions}
\label{sec:problem-formulation-assumptions}

The objective of the proposed RL setup is to optimize the efficiency of wind turbines and the sustainability of DC operation by making use of curtailment energy, that is, wind energy unwanted by the grid operator to maintain system stability or to avoid local transmission line congestion. To this end, the simulation environment aims to complete the compute job in a reasonable time, while maximizing the usage of wind energy and, especially, the usage of curtailment energy. \\

\subsection{Problem Formulation}

\begin{figure}[!ht]
    \centering
    \includegraphics[width=\linewidth]{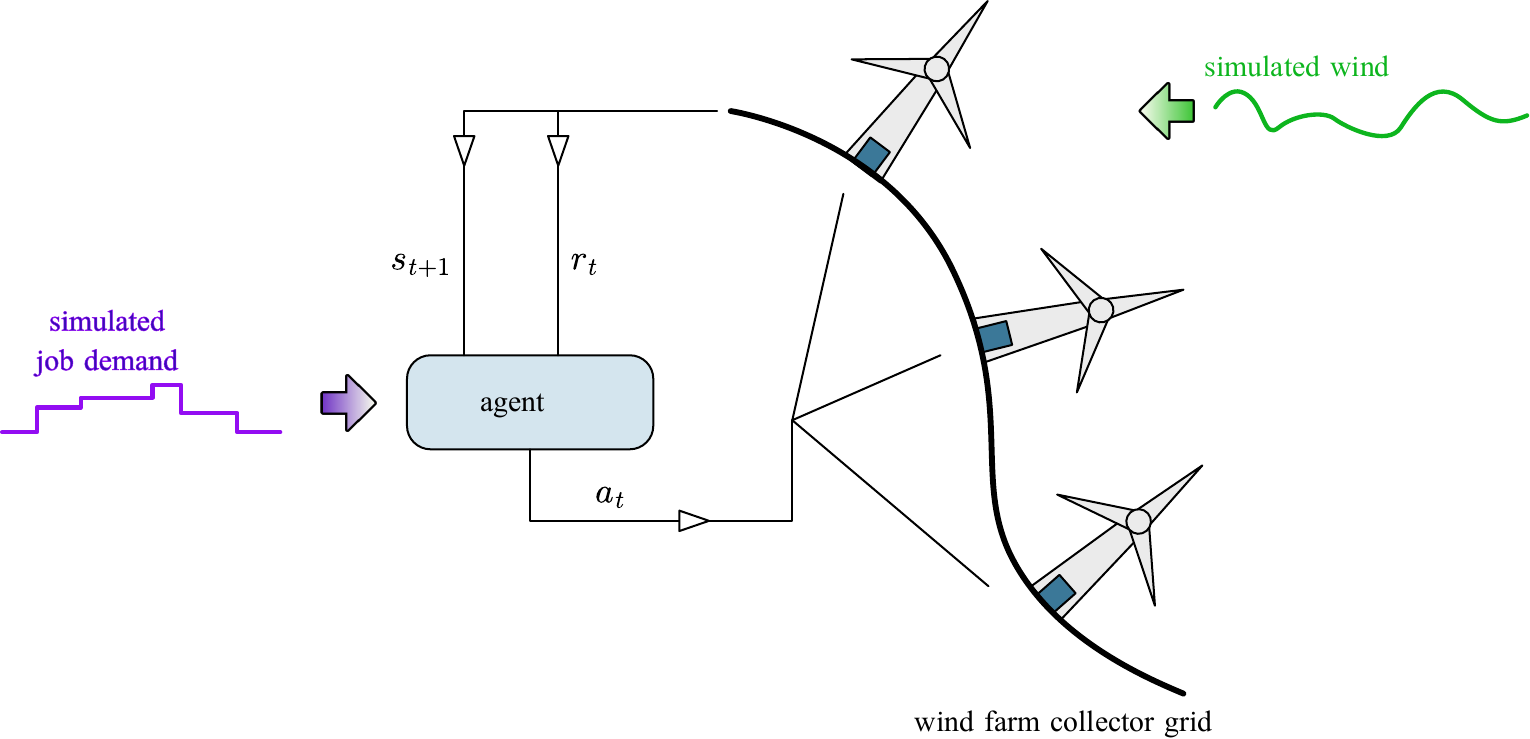}
    \caption{Simulation environment.}
    \label{fig:RL_Simulation_Environment}
\end{figure}

In our scenario, the basic idea introduced in Section \ref{sec:background-motivation} translates as follows (see Figure \ref{fig:RL_Simulation_Environment}): We consider a wind farm in which wind turbines house HPC-DCs that execute incoming jobs. It is connected to an external grid via a substation, where wind power can be injected into the grid, or power can be drawn from the external grid for DC supply.
The environment comprises wind farm data, such as the number of wind turbines that house an HPC-DC, resource utilization, and external signals, like wind forecast data and information about the external grid in terms of consumption costs, carbon intensity, etc. Furthermore, it contains information about compute jobs, as we need to track the progress regarding job execution. Based on the collected information, the agent needs to take some action to reduce the compute load left. To this end, it can specify the amount of compute power/resources used in each HPC-DC in each time step.

\subsection{General Idea and Assumptions}

The usage of wind power for powering HPC servers shall be rewarded. We assume a curtailment threshold that represents the wind energy share that must be fed into the grid. Any generation above this threshold would otherwise be curtailed and is therefore treated as curtailment energy that can be used locally for DC supply. In each time step, this free-energy component is given by $\max(0, \text{wind} - \text{curtailment threshold})$. If the wind is low or the compute demand exceeds this free share, the remaining demand is covered by external-grid electricity, called \textit{gray electricity}, and such gray electricity consumption is penalized more strongly when electricity prices (or, equivalently, carbon intensity of gray electricity) are high.

Finally, besides sustainability criteria, the most relevant criterion is concerned with performance: The agent shall be rewarded for reducing the total compute load reasonably fast, and finish the incoming jobs before some predetermined user-specified deadline. The objective is to find a proper trade-off between the discussed sustainability and performance criteria. Since we address HPC applications, we strive for a more performance-prone configuration of the simulation environment, i.e., the agent should learn to always complete the job in time. 

We will implement a simulation environment with a finite episode spanning one day, in the following called fixed-day simulation environment. We propose a minimal configuration of the simulation environment, which serves as an introductory example concerning the problem formulation. 

%% file: 5_Fixed_Day_Minimal.tex
\section{Fixed-Day Simulation Environment And Minimal Configuration}
\label{sec:fixed-day-minimal}

\paragraph*{Environment Overview}
In the fixed-day simulation, an episode spans one day and is treated as a finite-horizon planning problem. Each simulation step covers a time span of $\Delta t = 5$ minutes, which yields $n_T = t_e/\Delta t = 288$ control steps per episode. Here, $t_e = 1440$ minutes is the end time of one day. Time steps $t$ are normalized such that a change of 1.0 represents 24 hours. A single global job is scheduled whose computation load decreases from 1.0 to 0.0 over the course of the day. In this stage, we do not specify any relationships of this value with actual compute jobs.

We denote the number of simulated wind turbines by $N$. Furthermore, we assume that each wind turbine houses a DC.

At each time step, the environment uses manually crafted functions to generate wind and grid price signals. The curtailment threshold is kept constant at $ \theta = 0.4$, that is, 40\% of the generated wind power needs to be injected into the grid. While this fixed value is not meant to be fully realistic, it provides a stable benchmark setting to test whether controllers react to signals as intended in a proof-of-concept setup. The framework itself allows dynamic thresholds, but they are intentionally excluded in the present minimal benchmark. The remaining power can be used for server power supply for free. If power demand exceeds available free wind power, the difference needs to be compensated by the external grid, i.e., gray electricity is drawn from the grid connection point.

\begin{figure}[!ht]
    \centering
    \includegraphics[width=\linewidth]{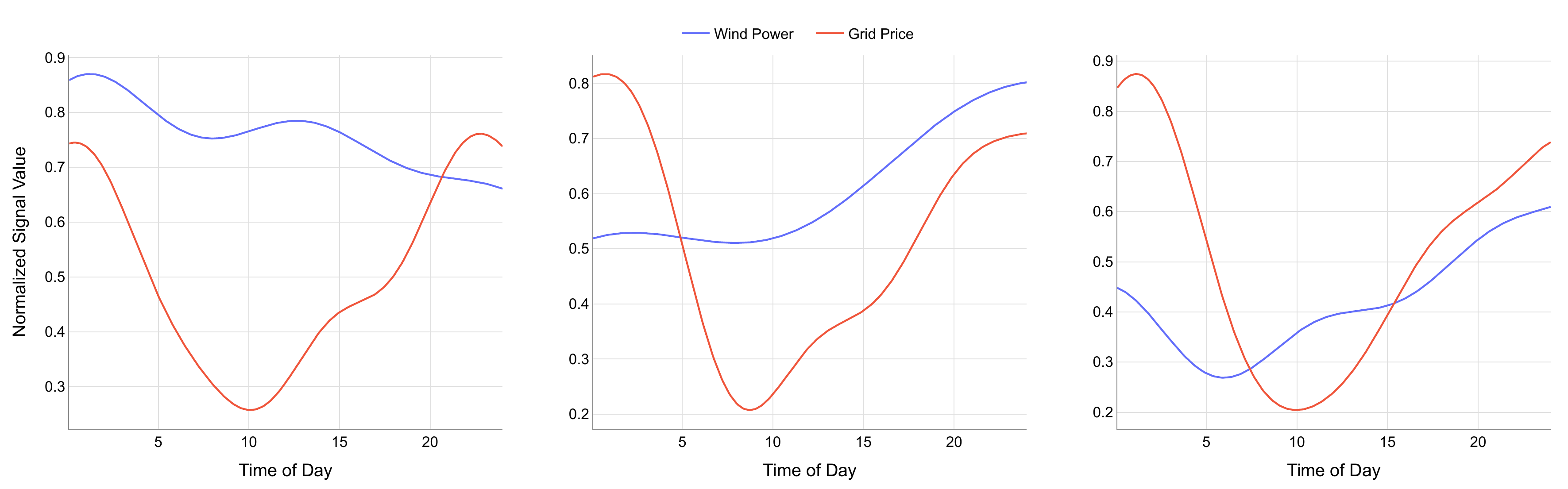}
    \caption{Example wind-power and grid price profiles over full days under the minimal configuration.}
    \label{fig:FixedDay_Signal_Generation}
\end{figure}

\paragraph*{Signal Generation}
The wind and grid price signals are generated synthetically for controllability and reproducibility. Wind is represented by a combination of smooth periodic components with randomized amplitudes/phases and a day-scale modulation. The grid price follows a smooth daily profile with lower values around midday and higher values near the beginning and end of the day, plus mild periodic perturbations. By mapping heterogeneous signals to comparable normalized ranges and relative scales, the same framework can be extended to additional signal types while keeping the control formulation consistent. Figure \ref{fig:FixedDay_Signal_Generation} shows example plots of these signals.

\subsection{State Vector}
\label{subsec:state-vector-minimal}
Let $t_k \in [0,1]$, $k \in \{1, \ldots, n_T\}$, be the current time step, where $n_T \in \mathbb{N}$ is the number of total time steps considered per episode, $t_1 = 0$, $t_{n_T} = 1$, and $t_{k+1} - t_k = \Delta t$. In particular, the time step $t_k$ represents the continuous time interval $[t_k,t_{k+1}]$.

For the minimal fixed-day setup, we use a single state vector based on first- and second-order discrete difference quotients. The observable state at time step $t_k$ is
\begin{align*}
\label{eq:state-vec-fixed-day-minimal}
    S_{t_k}
    = \big[ c_{t_k}, g_{t_k}, D_g^{(1)}(t_k), D_g^{(2)}(t_k), \theta_{t_k},
    \Xi_{t_k,1}^{\top}, \ldots, \Xi_{t_k,N}^{\top}, t_k \big]^{\top}
    \in \mathbb{R}^{4N+6},
\end{align*}
with
\begin{align*}
    \Xi_{t_k,i}
    := \big[w_{t_k,i}, D_{w,i}^{(1)}(t_k), D_{w,i}^{(2)}(t_k), w_{t_k,i}^{\mathrm{free}}\big]^{\top}
    \in \mathbb{R}^{4}, \qquad i=1,\ldots,N,
\end{align*}
where, with $\Delta t = 5$ minutes,
\begin{align*}
    D_g^{(1)}(t_k) &= \frac{g_{t_k} - g_{t_{k-1}}}{\Delta t},
    &
    D_g^{(2)}(t_k) &= \frac{g_{t_k} - 2g_{t_{k-1}} + g_{t_{k-2}}}{\Delta t^2},\\
    D_{w,i}^{(1)}(t_k) &= \frac{w_{t_k,i} - w_{t_{k-1},i}}{\Delta t},
    &
    D_{w,i}^{(2)}(t_k) &= \frac{w_{t_k,i} - 2w_{t_{k-1},i} + w_{t_{k-2},i}}{\Delta t^2}, \qquad i=1,\ldots,N.
\end{align*}
The lagged values at $k=-1$ and $k=-2$ are generated together with the synthetic wind and grid price signals and used directly in the difference quotients.

It consists of the following fields:
\begin{itemize}
 \item The remaining computation work for the day, $c_{t_k} \in [0,1]$, with $c_{t_0} = 1$;
  \item The current grid price $g_{t_k} \in [0,1]$;
  \item The first-order grid price difference quotient $D_g^{(1)}(t_k)$;
  \item The second-order grid price difference quotient $D_g^{(2)}(t_k)$;
  \item The current curtailment energy threshold, $\theta_{t_k} \in [0,1]$, which is assumed to be constant in the fixed-day simulation, i.e., $\theta_{t_k} = \theta$;
  \item For each wind turbine ($i=1,\dots, N$):
  \begin{itemize}
    \item The current wind power value $w_{t_k,i} \in [0,1]$;
    \item The first-order wind-power difference quotient $D_{w,i}^{(1)}(t_k)$;
    \item The second-order wind-power difference quotient $D_{w,i}^{(2)}(t_k)$;
    \item The available free wind power, 
    \begin{align*}
        w_{t_k, i}^{\mathrm{free}} = \max\left(0, w_{t_k, i} - \theta\right) \in [0, 1- \theta];
    \end{align*}
  \end{itemize}
  \item The current normalized time $t_k \in [0,1]$ (indicating how much time is left to do the computation work).
\end{itemize}

Including $w_{t_k,i}^{\mathrm{free}}$ explicitly in the state is a deliberate design choice to make the amount of immediately usable curtailment energy directly observable to the agent, especially at early time steps. This improves observability, but by itself does not remove the delayed-feedback challenge.

In this work, we assume a one-to-one relation between most of the introduced variables (e.g., the wind signal equals the generated wind power).

By \emph{minimal configuration}, we explicitly mean the single-site baseline with $N=1$: one wind turbine with one co-located HPC-DC. Consequently, the agent controls one utilization value per time step. Unless otherwise stated, all plots shown in this section refer to this minimal configuration ($N=1$).

Even in this minimal setting, the learning problem already exhibits delayed feedback: in the baseline reward design, completion shortfall is penalized only at the end of the episode. Consequently, early utilization decisions are linked to a terminal signal with a long temporal gap, which creates a non-trivial credit-assignment challenge.

\subsection{Action Vector}
\label{subsec:action-vector-minimal}
\noindent Based on the observed state, the agent learns and picks a suitable action $A_{t_k}$ at each time step $t_k$. Policy outputs are continuous values in $[-1,1]^N$ and are mapped affinely to utilization values in $[0,1]^N$. More precisely, with raw policy output
\begin{equation*}
    \widetilde{A}_{t_k} = (\tilde{a}_{t_k,1},\ldots,\tilde{a}_{t_k,N})\in[-1,1]^N,
\end{equation*}
the executed action is
\begin{equation*}
    A_{t_k} = (u_{t_k,1}, \ldots, u_{t_k, N}), \qquad
    u_{t_k,i} = \frac{\tilde{a}_{t_k,i} + 1}{2}\in[0,1].
\end{equation*}
In particular, we allow that during execution, there might be time steps where no calculations are carried out due to poor wind conditions or high gray electricity consumption costs. In other words, checkpointing is a feasible option and job execution can be proceeded at a later point in time. For the baseline configuration ($N=1$), the action is therefore a single utilization scalar per step.

\subsection{Reward Design}
\label{subsec:reward-design-minimal}
Assuming that per time step, the distributed HPC-DC can reduce the computation work by at most 1\% (if operated at full capacity), this allows us to determine the suggested cumulated compute power, which is bounded by the computation work left at $t_k$:
\begin{align*}
    P_{t_k}^{\mathrm{comp}} =  \min \left( c_{t_k}, \frac{1}{N} \sum_{i = 1}^{N} u_{t_k, i} \cdot 0.01 \right) \in [0, 0.01].
\end{align*}
The computation work left at $t_{k+1}$ is given by
\begin{equation}
    c_{t_{k+1}} = \max \left(0,  c_{t_k} -P_{t_k}^{\mathrm{comp}}\right).
    \label{eq:remaining-work}
\end{equation}

In the present scenario, we assume that not all the generated wind power is actually usable to power the HPC-DCs. In fact, only if more than $\theta = 40~\%$ of the nominal capacity of the respective wind turbine is generated, the HPC servers run with the energy generated on site. This energy surplus corresponds to the curtailment energy and is free to use. The state values $w_{t_k, i}^{\mathrm{free}}$ constitute the total free wind power:

\begin{align*}
    P_{t_k}^{\mathrm{free}} = \sum_{i = 1}^{N} w_{t_k,i}^{\mathrm{free}} = \sum_{i=1}^{N} \max \left( 0, w_{t_k, i} - \theta \right).
\end{align*}
The reward at time $t_k$ is given by
\begin{equation*}
    R_{t_k} = -\psi_{\beta,\delta}\!\left(P_{t_k}^{\mathrm{comp}} - P_{t_k}^{\mathrm{free}}\right)\cdot g_{t_k},
\end{equation*}
with
\begin{equation*}
    \psi_{\beta,\delta}(x) := \frac{N}{100\beta}\log\!\left(1+\mathrm{e}^{\beta\left(\frac{100}{N}x-\delta\right)}\right), \qquad \beta>0,\ \delta \ge 0.
\end{equation*}
This definition explicitly includes the normalization of compute power ($\tfrac{100}{N}$) before the softplus and the corresponding denormalization ($\tfrac{N}{100}$) of the result. The shifted softplus term is a smooth approximation of the positive mismatch between compute demand and free wind power (i.e. $\max(0,P_{t_k}^{\mathrm{comp}} - P_{t_k}^{\mathrm{free}})$) and avoids hard kinks in the reward landscape. In our experiments, we set $\beta=700$ and $\delta=0.006$. Hence, the agent is punished if more power is used than needed to perform computations from free wind power at $t_k$. The penalty for drawing gray electricity is weighted by the grid price $g_{t_k}$.   \\ [1ex]

In the baseline reward design (without shaping), delayed completion is penalized only at the end of the episode. Concretely, we check whether $c_{t_{n_T}} = 0$ holds and define
\begin{equation*}
    R^{\mathrm{end}} = - c_{t_{n_T}}.
\end{equation*}
This corresponds to a (negative) final-state reward. 

Overall, the cumulated reward for each episode, which has to be maximized by the agent, is given by
\begin{align}
    R = \sum_{k = 1}^{n_T} R_{t_k} + R^{\mathrm{end}}.
    \label{eq:return-with-terminal}
\end{align}
If $c_{t_{k^{\ast}}} = 0$ for some $k^{\ast} \leq n_T-1$ (see \eqref{eq:remaining-work}), then the episode ends early, and the total reward amounts to
\begin{equation*}
    R = \sum_{k= 1}^{k^{\ast}} R_{t_k}.
\end{equation*}

As, in general, DC power supply using only free wind power is not sufficient to complete the job by the end of an episode, the agent needs to learn how to efficiently manage resource utilization. The objective is to complete the compute job in time (to avoid final-state punishments), while carefully choosing favorable periods, that is, periods with high wind values and/or low energy consumption prices, for carrying out calculations. 

In this setup, learning is non-trivial: under-allocation of HPC resources, when curtailment energy is available, is penalized mainly through the terminal term $R^{\mathrm{end}}$, so the adverse consequence of early low-utilization actions is only visible at the end of the episode. This induces a delayed-feedback credit-assignment problem, which is strongest near the beginning of an episode where many decisions lie between action and terminal signal. 

To address this delayed-feedback issue, we investigate two mitigation paths in the following order: first optimization-based IL, then potential-based RS.

\subsection{Optimization-based Imitation Learning}
\label{subsec:optimization-il-minimal}
Since an episode in this simulation spans an entire day and the simulator can generate all daily signals and dynamics beforehand, it is possible to run an offline optimization algorithm to find optimal or near-optimal solutions for each day. The optimization problems are solved with JuMP\footnote{\url{https://github.com/jump-dev/JuMP.jl}} using Ipopt\footnote{\url{https://github.com/coin-or/Ipopt}}. This a-priori information is used for benchmark construction and expert-trajectory generation, but it is not available to the RL policy during online control. At each decision step, the agent only receives the current state. We have first introduced this optimization layer as a reference method to provide a strong algorithm-independent benchmark for policy evaluation. After diagnosing the delayed-feedback credit-assignment issue in pure RL training, especially the missing early use of free wind energy, we then have used optimization trajectories as supervision signals for IL as a first targeted mitigation strategy. In practice, one may generate whole-day episodes, run an offline optimization of the corresponding load distribution task over these episodes, and then feed the resulting trajectories (or expert actions) into the agent’s memory buffer. The agent can subsequently run its update routine using both its own experience and these optimal trajectories. \\ [1ex]
 
 Denote by
\begin{equation*}
    U_k = (u_{k,1}, u_{k,2}, \ldots, u_{k,N}) \in [0,1]^{N}, \quad k \in \{1, \ldots, n_T\},
\end{equation*}
the vector capturing the utilization of each DC at time step $k$. The vector 
\begin{equation*}
    U = (U_1, \ldots, U_{n_T}) \in [0,1]^{N\cdot n_T}
\end{equation*}
contains the decision variables of the present optimization problem.

For the offline optimizer, we collect the full-day environmental forecasts in the vectors
\begin{align*}
     G &= ( G_1, \ldots, G_{n_T} ) \in [0, 1]^{n_T} \quad \text{and} \\
    W &= ( W_1, \ldots, W_{n_T}) \in [0, 1]^{N \cdot n_T},
\end{align*}
respectively. The objective is to find the optimal load distribution, depending on the environmental parameters, to maximize the cumulated reward. It is given by
\begin{equation*}
    R := R_{G,W} \colon [0,1]^{N \cdot n_T}  \to \mathbb{R}_-,  \quad U \mapsto  R(U) := \sum_{k=1}^{n_T} R_k (U_k),
\end{equation*}
where $R_k: = R_{G_k, W_k}  \colon [0,1]^N \to \mathbb{R}_-$ is the reward function parameterized by the environmental data at time step $k$. The reward function is designed as in Section~\ref{subsec:reward-design-minimal}.

It is required that the agent distributes the total computation load in time. Recall that per time step, at most 1\% of the total load can be computed. Thus, we impose the following constraint:
\begin{align*}
    \frac{1}{N}\sum_{k=1}^{n_T} \sum_{i=1}^{N} u_{k,i} \cdot 0.01 = 1 \quad \text{or, equivalently,} \quad     \sum_{k=1}^{n_T} \sum_{i=1}^{N} u_{k,i}  = 100N.
\end{align*}    

Altogether, the Discrete Optimization Problem (DOP) can be formalized as follows:
\begin{align}
        &\text{maximize}  \; R(U) = \sum_{k=1}^{n_T} R_k (U_k) \nonumber \\
        &\text{with respect to}\quad  U =(U_1, \ldots, U_{n_T}) \in [0,1]^{N\cdot n_T}   \label{DOP} \tag{DOP} \\
        &\text{subject to}  \quad  \sum_{k=1}^{n_T} \sum_{i=1}^{N} u_{k,i}  = 100N. \nonumber
\end{align}
By solving \eqref{DOP}, we obtain a reference load distribution trajectory based on expert actions for a selection of episodes.  \\ [1ex]

To prefer earlier allocation, we introduce an additional objective
\begin{equation*}
    F(U) := \frac{1}{100N \cdot n_T}\sum_{k=1}^{n_T} (n_T-k+1)\sum_{i=1}^{N}u_{k,i},
\end{equation*}
which assigns larger weights to earlier time steps. We then solve a scalarized multi-objective optimization problem with weighted aggregation:
\begin{align}
        &\text{maximize} \; J(U) := R(U) + \lambda F(U) \nonumber \\
        &\text{with respect to}\quad  U =(U_1, \ldots, U_{n_T}) \in [0,1]^{N\cdot n_T} \label{DOP_weighted} \tag{DOP-W} \\
        &\text{subject to}  \quad  \sum_{k=1}^{n_T} \sum_{i=1}^{N} u_{k,i}  = 100N, \nonumber
\end{align}
where $\lambda \ge 0$ controls the trade-off between primary reward maximization and early allocation preference.

\begin{figure}[!ht]
    \centering
    \includegraphics[width=\linewidth]{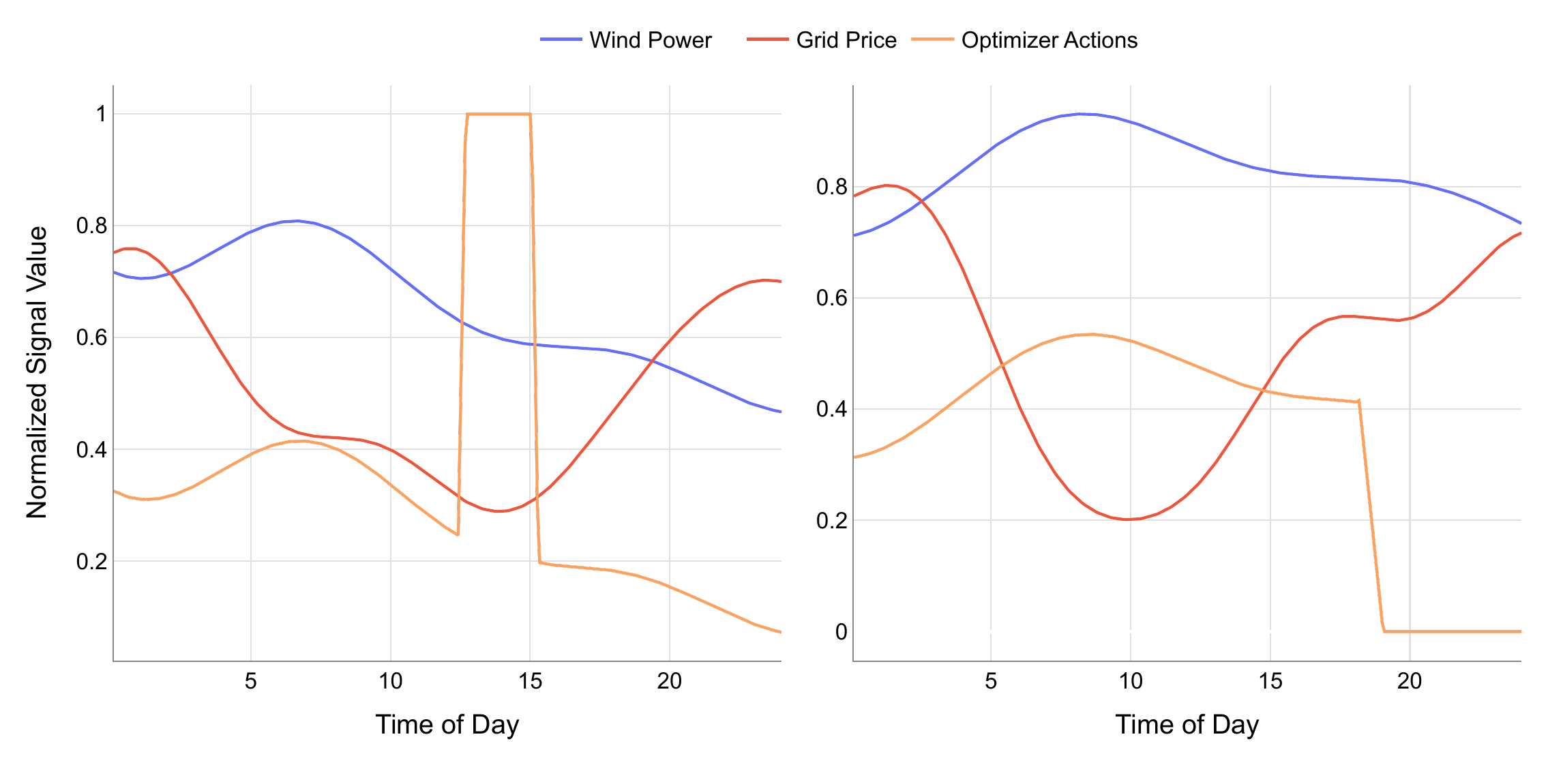}
    \caption{Results obtained from the optimization algorithm on fixed-day episodes in the minimal configuration.}
    \label{fig:FixedDay_Optimal}
\end{figure}

Figure \ref{fig:FixedDay_Optimal} illustrates typical results obtained from the optimization algorithm applied to a fixed-day episode. Here, we set $N=1$ and $n_T= 288$. In the first example (on the left), the servers should run at full capacity when grid consumption prices (red) are lowest, as there is not enough free wind power to complete the compute job in time. In the second example (on the right), the optimizer output suggests using all of the free wind power: the utilization trajectory (green) is essentially the same as the wind power trajectory (blue) shifted by the curtailment threshold $\theta = 0.4$.

This allows us to assess intermediate results and improve the overall performance of the RL-agent via IL.
During RL training, imitation updates are not executed only once as pretraining; instead, they are injected repeatedly into the standard update loop by alternating IL updates and online rollout updates. We precompute trajectory sets to keep ablation studies consistent.

For IL configurations, the update cadence is chosen such that IL updates account for approximately one quarter of all parameter updates.

The expert corpus used for IL consists of trajectories from 2000 randomly generated days. For PPO-based IL, we follow the guided-exploration idea of PPO+D \cite{libardi2021}. In this setting, the log-probability assignment for expert actions is not uniquely prescribed. In our experiments, we use a fixed surrogate value of $0.1$ for expert-action log probabilities. This value is treated as an IL hyperparameter and was found to work well in the present setup. A broader sensitivity study over IL hyperparameters (e.g., surrogate log-probabilities and IL update cadence) is beyond scope here, because the joint search space grows quickly in dimensionality. Because the expert corpus is large, PPO-IL updates are performed on random windows of 8000 time steps, matching the standard PPO update size used in our training loop.

For standard SAC, IL is integrated by sampling batches from the expert corpus and applying the unchanged off-policy SAC update routine. For the SAC variant with an additional on-policy routine, IL uses that on-policy routine together with the same random-window strategy (8000 time steps) as in PPO. The motivation is that, especially without RS, action quality can only be judged reliably from whole-episode context because early actions are coupled to delayed completion signals. Hence, the additional on-policy routine explicitly reuses coherent recent trajectories and proved more effective than pure off-policy SAC in our setting. A detailed description of this SAC on-policy routine is provided in~\ref{app:sac_onpolicy_update}.

Interleaving these expert-driven updates slows training progress per episode because the agent is repeatedly steered away from its own on-trajectory experience toward externally generated trajectories. Therefore, IL configurations are trained for substantially more episodes than their non-IL counterparts.

\subsection{Potential-based Reward Shaping}
\label{subsec:pbrs-minimal}
As a second mitigation strategy next to IL, we study potential-based RS in the sense of \cite{Ng1999}:
\begin{equation*}
    \widetilde{R}_{t_k} = R_{t_k} + \eta \left(\gamma \Phi(c_{t_{k+1}})-\Phi(c_{t_k})\right),
\end{equation*}
with shaping weight $\eta \ge 0$ and potential function $\Phi(c)=-c$. The potential term follows the RS construction and is policy-invariant under the standard assumptions of \cite{Ng1999}. In our setting, this term redistributes completion-related feedback from the end of an episode to intermediate steps. As a consequence, the isolated end-of-episode penalty signal does not appear in the same way in shaped runs (compare with \eqref{eq:return-with-terminal}). Its effect is absorbed into the shaped per-step rewards over the trajectory. We therefore evaluate shaped runs via
\begin{equation*}
    \widetilde{R} = \sum_{k = 1}^{n_T} \widetilde{R}_{t_k},
\end{equation*}
instead of relying on a separate terminal penalty term. This transforms the sparse terminal-punishment setup into a denser temporal allocation problem with largely positive progress feedback. However, the core challenge remains: the agent still has to infer horizon effects and long-range action consequences to schedule load strategically over the day.

\subsection{Agent Training Results}
In this section, we compare different realizations of the simulation environment and training pipeline using the fixed derivative-based state vector from Subsection~\ref{subsec:state-vector-minimal}. Concretely, we compare training with and without IL, both with and without potential-based RS.

All results reported in this subsection are obtained strictly under the minimal configuration ($N=1$), even though the general setup supports more complex multi-site variants. We keep $N=1$ here on purpose because this setting provides a controlled system-level benchmark in which differences between training strategies can be compared without additional structural confounders.

The training pipeline is evaluated in a reproducible ablation setup across multiple algorithms. We use disjoint validation and test sets, each comprising 200 days that are not part of the training data. For each algorithmic configuration, six independent training runs are executed. Model selection is performed on the validation set by choosing the best run, and only this selected policy is evaluated on the test set.

This setup makes it possible to separate two effects in a controlled way: optimization-informed supervision (IL vs no IL) and reward-design variants (with/without potential-based RS).

\begin{figure}[!ht]
    \centering
    \includegraphics[width=\linewidth]{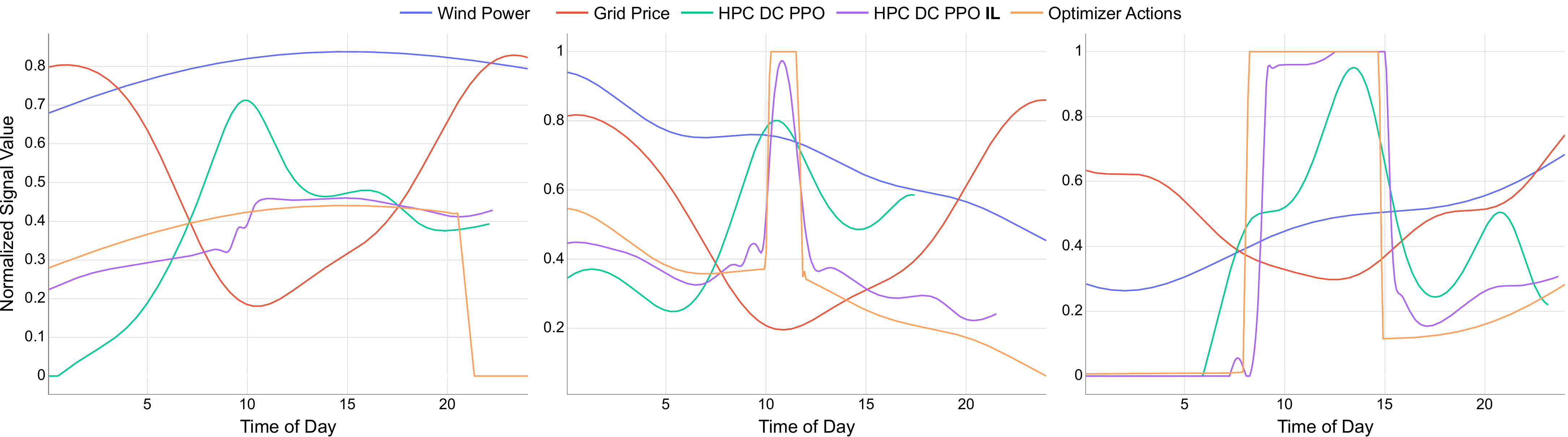}
    \caption{Results of fully trained PPO agents (with and without IL) and the corresponding optimizer output for three days of simulation under the minimal setup.}
    \label{fig:PPOresults}
\end{figure}

Figure \ref{fig:PPOresults} shows that IL substantially improves early-day use of free wind energy. The PPO policy with IL follows the optimizer trajectory much more closely, especially at the beginning of the day where delayed credit assignment is strongest. The PPO policy without IL tends to allocate too conservatively in these early periods and compensates later with less favorable allocations.

\begin{figure}[!ht]
    \centering
    \includegraphics[width=\linewidth]{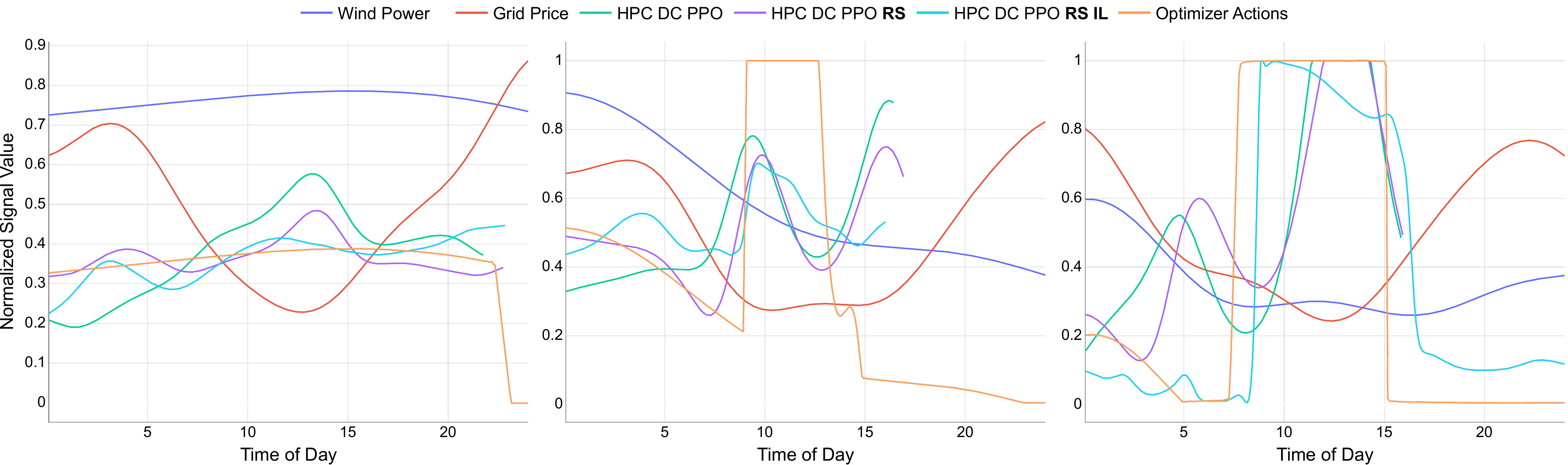}
    \caption{Results of fully trained PPO agents (without IL and RS, with RS and with RS and IL) and the corresponding optimizer output for three days of simulation under the minimal setup.}
    \label{fig:PPO_RS_results}
\end{figure}

\begin{figure}[!ht]
    \centering
    \includegraphics[width=\linewidth]{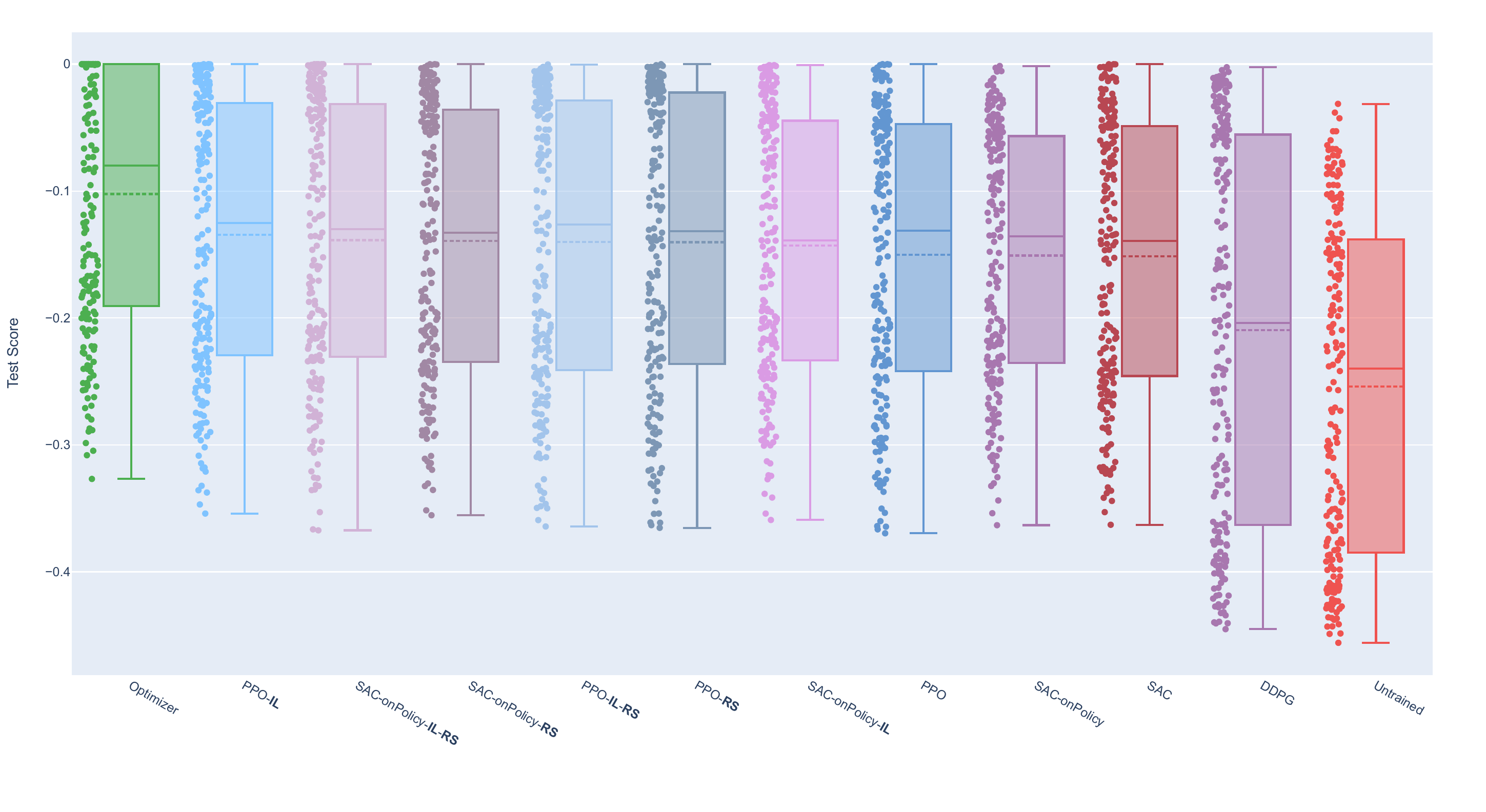}
    \caption{Box-plot comparison of algorithm performance on the 200-day test set under the minimal setup, including optimizer and untrained-agent baselines.}
    \label{fig:AlgorithmPerformance}
\end{figure}

Figure \ref{fig:PPO_RS_results} indicates that RS improves early-day utilization of free wind energy. At the same time, RS-only runs show less stable control strategies over the full day horizon. Combining RS with IL yields a mixed behavior, but not a strictly better one across the complete trajectory. This interpretation is consistent with the aggregate ranking in Table~\ref{tab:algorithm-performance-table} and the distributional comparison in Figure~\ref{fig:AlgorithmPerformance}.

Figure \ref{fig:AlgorithmPerformance} reports box plots for a selected subset of tested configurations under the minimal setup; for complete mean scores across all combinations, see Table~\ref{tab:algorithm-performance-table}. In each box, the dashed line denotes the mean and the solid line denotes the median. The point cloud shown next to each box corresponds to the individual scores of all 200 test-set days for the respective algorithm configuration. In addition to learned agents, we include the optimizer performance on the test set and the performance of an untrained agent.
For readability, Figure \ref{fig:AlgorithmPerformance} shows only the no-IL/no-RS variants for DDPG and standard SAC. Their additional IL/RS variants do not show strong gains in this fixed-day minimal setting, suggesting that these mitigation techniques are more effective here when coupled with on-policy-style procedures.
For PPO and SAC-on-policy, all four IL/RS combinations are shown. For both algorithms, IL and RS each act as effective mitigation strategies, while combining both is not strictly better than selecting one of them.
When interpreting the gap to the optimizer, one should keep the information structure in mind: the optimizer solves a full-day offline planning problem with access to the complete day trajectory, whereas the RL policy acts online and must decide step by step from current observations without future-day realizations. The optimizer is therefore a strong upper reference, not a level that online RL is expected to match exactly.
The untrained agent serves as a lower baseline and represents a comparatively simple control behavior. With initialization that yields raw actions mostly around zero, it often still completes the daily target and frequently operates in comparatively favorable periods. During training, agents first move away from this trivial behavior and can temporarily become markedly worse because delayed terminal penalties appear unexpectedly. Only after longer training do they recover and eventually surpass the untrained baseline with more effective control policies.

Table~\ref{tab:algorithm-performance-table} provides the complete mean-score overview across all IL/RS combinations, including those not visualized in Figure~\ref{fig:AlgorithmPerformance}.
Complementary per-episode means for curtailment energy usage, gray electricity consumption, deadline-violation rate, and remaining deadline load are reported in~\ref{app:per_episode_metrics}, Table~\ref{tab:per-episode-metrics-all-configs}. As expected, the optimizer shows the highest curtailment energy usage, while the untrained agent shows the lowest. Across learned policies, strong mean score does not always coincide with maximal curtailment energy usage: for example, all DDPG configurations achieve comparatively high curtailment energy means despite weak score ranking in Table~\ref{tab:algorithm-performance-table}. At the same time, PPO (IL, with RS) is among the strongest score-competitive settings and also remains high in curtailment energy usage. This pattern suggests that score differences are not driven by curtailment energy uptake alone. They are also linked to how much and when gray electricity is used, i.e., whether gray electricity consumption occurs at less favorable price periods. Finally, while some configurations exhibit non-zero deadline-violation rates, the corresponding mean remaining load is very small (all values below $0.005$), so these violations are practically minor in this setup because only negligible residual work is billed at the end-of-episode maximum grid price.

\begin{table}[!ht]
    \centering
    \caption{Mean algorithm performance on the 200-day test set in the minimal configuration ($N=1$), grouped by IL/RS combinations. Lower values (in magnitude) are better. The bold entry marks the best-performing learned algorithm configuration.}
    \label{tab:algorithm-performance-table}
    \small
    \begin{tabular}{lcccc}
        \hline
        Algorithm & no IL, no RS & IL, no RS & no IL, with RS & IL, with RS \\
        \hline
        Optimizer (offline reference) & -0.102 & -- & -- & -- \\
        PPO & -0.150 & \textbf{-0.135} & -0.140 & -0.140 \\
        SAC-on-policy & -0.151 & -0.143 & -0.139 & -0.139 \\
        SAC & -0.151 & -0.166 & -0.152 & -0.156 \\
        DDPG & -0.210 & -0.207 & -0.211 & -0.212 \\
        Untrained agent & -0.254 & -- & -- & -- \\
        \hline
    \end{tabular}
\end{table}

%% file: 8_Conclusion.tex
\section{Conclusions and future work}
\label{sec:conclusion-future-work}

This paper introduced a fixed-day minimal RL environment for curtailment-aware HPC load allocation in wind-turbine-integrated DCs and compared multiple RL agents against optimization references. The results show a persistent delayed-feedback credit-assignment problem: without additional guidance, agents tend to underuse free wind energy early in the episode. We therefore investigated two complementary mitigation paths, optimization-based IL and potential-based RS. In the reported benchmark, both measures improve policy behavior in relevant settings, and PPO as well as the SAC variant with an additional on-policy update routine achieve the strongest empirical performance among the learned controllers.

At the same time, a non-zero gap to the optimization reference remains and is expected: the optimizer solves an offline full-day planning problem with complete future information, while RL policies act online under uncertainty and only observe the current state. Hence, the optimizer should be interpreted as an upper reference for online control rather than as a directly reachable target.

All experiments reported in this manuscript are available as open-source code at \url{https://github.com/janstenner/ESN4NW-RL-public}. The same repository also contains further experiments that are not part of this paper's scope, including runs with real data sources, prediction models in synthetic-data-in-the-loop settings, and continuous-time simulations with multiple job slots, categorical actions, and multi-agent support. Extending the present evaluation along these lines is a natural direction for future work.

%% file: A_Appendix_SAC_OnPolicy.tex
\section{On-Policy Update Routine for SAC}
\label{app:sac_onpolicy_update}

This appendix summarizes the on-policy SAC update routine used in the fixed-day experiments. The routine is applied periodically during rollout training and, for imitation learning, on randomly sampled windows from the expert corpus.

\begin{algorithm}[H]
\small
\DontPrintSemicolon
\caption{SAC On-Policy Update (rollout and IL mode)}
\label{alg:sac_onpolicy_update}
\SetKwInOut{Input}{Input}
\SetKwInOut{Output}{Output}
\Input{replay trajectory $\mathcal{T}$, optional expert trajectory $\mathcal{D}_{\mathrm{exp}}$, mode $m\in\{\text{rollout},\text{IL}\}$, actor $\pi_{\theta}$, critics $Q_{\phi_1},Q_{\phi_2}$, target critics $\bar Q_{\phi_1},\bar Q_{\phi_2}$, coefficients $\gamma,\tau,\alpha,\lambda_{\mathrm{targets}}$, target entropy $\mathcal{H}_{\mathrm{target}}$, antithetic samples $K$, epochs $E$, mini-batches $M$, window size $n_{\mathrm{on}}$}
\Output{updated parameters $\theta,\phi_1,\phi_2,\alpha$ and target critics}

\If{$m=\text{rollout}$}{
  use the most recent $n_{\mathrm{on}}$ transitions from the replay trajectory (which is large and also used for off-policy updates)\;
}
\Else{
  set $n\leftarrow \min(n_{\mathrm{on}},|\mathcal{D}_{\mathrm{exp}}|)$ and sample a random contiguous expert window of length $n$\;
}

\tcp{Form transition window $\mathcal{D}=\{(s_t,a_t,r_t,d_t^{\mathrm{ter}},d_t^{\mathrm{trun}},s_{t+1})\}$}
\ForEach{$s_{t+1}$ in $\mathcal{D}$}{
  draw $K$ antithetic action pairs $(a_t^{+},a_t^{-})\sim\pi_{\theta}(\cdot|s_{t+1})$\;
  \[
  \hat v_{t+1}=\frac{1}{2K}\sum_{k=1}^{K}\!\left[\begin{aligned}
  \phantom{+}\min_j \bar Q_{\phi_j}(s_{t+1},a_{t,k}^{+}) &-\alpha\log\pi_{\theta}(a_{t,k}^{+}|s_{t+1}) \\
  +\min_j \bar Q_{\phi_j}(s_{t+1},a_{t,k}^{-}) &-\alpha\log\pi_{\theta}(a_{t,k}^{-}|s_{t+1})
  \end{aligned}
  \right],
  \]
  \[
  \widehat{\log\pi}_{t+1}=\frac{1}{2K}\sum_{k=1}^{K}\left(
  \log\pi_{\theta}(a_{t,k}^{+}|s_{t+1})+\log\pi_{\theta}(a_{t,k}^{-}|s_{t+1})
  \right).
  \]
}

compute temporal-difference (TD)($\lambda$) targets $G_t^\lambda$ from $(r_t,d_t^{\mathrm{ter}},d_t^{\mathrm{trun}},\hat v_{t+1})$ using the standard backward recursion\;

flatten the window into $n$ training samples\;

\ForEach{epoch $e\in\{1,\ldots,E\}$}{
  shuffle and split the $n$ samples into $M$ mini-batches\;
  \ForEach{mini-batch $\mathcal{B}$}{
    update each critic by minimizing $\mathcal{L}_{Q_j}=\mathbb{E}_{(s,a,G^\lambda)\in\mathcal{B}}[(Q_{\phi_j}(s,a)-G^\lambda)^2]$, $j\in\{1,2\}$\;
    polyak-update target critics: $\bar\phi_j \leftarrow (1-\tau)\bar\phi_j+\tau\phi_j$\;
    update actor by minimizing $\mathcal{L}_{\pi}=\alpha\,\mathbb{E}_{s\in\mathcal{B}}[\log\pi_{\theta}(\tilde a|s)]-\mathbb{E}_{s\in\mathcal{B}}[\min_j Q_{\phi_j}(s,\tilde a)]$, $\tilde a\sim\pi_{\theta}(\cdot|s)$\;
  }
}

update $\log\alpha$ with $\mathcal{L}_{\alpha}=\exp(\log\alpha)\,\mathbb{E}[-\widehat{\log\pi}_{t+1}-\mathcal{H}_{\mathrm{target}}]$, clamp $\log\alpha$, and set $\alpha\leftarrow\exp(\log\alpha)$\;
\end{algorithm}

%% file: B_Appendix_PerEpisode_Metrics.tex
\section{Per-Episode Metric Overview (Test-Set Means)}
\label{app:per_episode_metrics}

Table~\ref{tab:per-episode-metrics-all-configs} reports the mean per-episode metrics over all 200 test-set days for all evaluated configurations in the fixed-day minimal setting.

\begin{table}[!ht]
    \centering
    \caption{Per-episode metric means on the 200-day test set for all configurations in the minimal setup ($N=1$). Abbreviations: CEU = Curtailment Energy Usage, GEC = Gray Electricity Consumption, DVR = Deadline Violation Rate, DCL = Deadline Compute Left.}
    \label{tab:per-episode-metrics-all-configs}
    \scriptsize
    \begin{tabular}{p{4.8cm}rrrr}
        \hline
        Configuration & CEU & GEC & DVR & DCL \\
        \hline
        Optimizer (offline reference) & 59.180035 & 40.819983 & 0.000 & 0.000000 \\
        PPO (IL, no RS) & 54.523760 & 45.424604 & 0.045 & 0.000516 \\
        PPO (IL, with RS) & 54.857986 & 44.708035 & 0.160 & 0.004340 \\
        PPO (no IL, no RS) & 51.244232 & 48.472179 & 0.085 & 0.002836 \\
        PPO (no IL, with RS) & 55.158860 & 44.584911 & 0.080 & 0.002562 \\
        SAC-on-policy (IL, no RS) & 51.199299 & 48.655650 & 0.155 & 0.001451 \\
        SAC-on-policy (IL, with RS) & 52.593179 & 47.271905 & 0.165 & 0.001349 \\
        SAC-on-policy (no IL, no RS) & 48.486110 & 51.336720 & 0.100 & 0.001772 \\
        SAC-on-policy (no IL, with RS) & 53.188812 & 46.718473 & 0.100 & 0.000927 \\
        SAC (IL, no RS) & 46.853925 & 52.898437 & 0.170 & 0.002477 \\
        SAC (IL, with RS) & 47.254048 & 52.432768 & 0.105 & 0.003132 \\
        SAC (no IL, no RS) & 48.188306 & 51.618100 & 0.105 & 0.001936 \\
        SAC (no IL, with RS) & 48.368513 & 51.614140 & 0.020 & 0.000174 \\
        DDPG (IL, no RS) & 51.840039 & 48.100017 & 0.060 & 0.000599 \\
        DDPG (IL, with RS) & 51.039568 & 48.960431 & 0.000 & 0.000000 \\
        DDPG (no IL, no RS) & 53.312510 & 46.687490 & 0.000 & 0.000000 \\
        DDPG (no IL, with RS) & 54.138025 & 45.861975 & 0.000 & 0.000000 \\
        Untrained agent & 40.670481 & 59.329520 & 0.000 & 0.000000 \\
        \hline
    \end{tabular}
\end{table}